\documentclass{article}
\usepackage{spconf,amsmath,graphicx}
\usepackage{booktabs}
\usepackage{hyperref}
\usepackage{adjustbox}
\usepackage{multirow}
\usepackage[table,xcdraw]{xcolor}
\usepackage{adjustbox}
\usepackage{bbding}

\title{Generative Speech Recognition Error Correction with Large Language Models and Task-Activating Prompting}

\name{%
\begin{tabular}{@{}c@{}} Chao-Han Huck Yang, Yile Gu, Yi-Chieh Liu, Shalini Ghosh, Ivan Bulyko, Andreas Stolcke
\end{tabular}}
\address{Amazon, USA }
\begin{document}

\maketitle

\sloppy
 
\begin{abstract}
%
We explore the ability of large language models (LLMs) to act as speech recognition post-processors that perform rescoring and error correction.
Our first focus is on instruction prompting to let LLMs perform these task without fine-tuning, for which we evaluate different prompting schemes, both zero- and few-shot in-context learning, and a novel ``task activation'' prompting method that combines causal instructions and demonstration to increase its context windows. 
Next, we show that rescoring only by
in-context learning with frozen LLMs achieves results that are competitive with rescoring by domain-tuned LMs, using a pretrained first-pass recognition system and rescoring output on two out-of-domain tasks (ATIS and WSJ).
By combining prompting techniques with fine-tuning we achieve error rates below the N-best oracle level, showcasing the generalization power of the LLMs.


%
%
\end{abstract}

\section{Introduction}
Large-scale language models (LLMs) have exhibited outstanding performance on downstream tasks by conditioning on input information, including task descriptions (e.g., performing mathematical calculations) or a limited number of input-output pairs obtained from training text (e.g., goal-oriented demonstrations). This new capability of task-specific inference from contextual information has been referred to as ``\textit{in-context learning}'' in Brown \textit{et al.}~\cite{brown2020language}. More specifically, the ability to learn in-context has been reported in previous studies~\cite{ min2022rethinking} of pretrained LLMs with over $100$B parameters trained with an unsupervised auto-regressive objective. Although recent advances in in-context learning have consistently demonstrated excellent performance on a wide range of tasks~\cite{wei2022finetuned}, there have been limited studies on the interaction or benefits of in-context learning on automatic speech recognition (ASR) tasks. As an example, contextual information~\cite{martinez2021attention} has been shown to play a vital role on ASR applications in complex domains, such as recognizing utterances referring to trending news.

One open question in the development of robust ASR applications is \textit{how} recent in-context learning frameworks can utilize their zero-shot learning capability to enhance ASR systems. Meanwhile, scaling ASR model sizes up to $10$B parameters~\cite{li2021scaling} by itself has not proven adequate for achieving high performance on challenging (e.g., conversational) speech tasks from domain-specific data. The challenge to obtain better generalization of neural ASR models has motivated proposals to incorporate external knowledge from textual data~\cite{chen2023hp}. For instance, one way to improve the RNN-transducer is to incorporate an external LM~\cite{le2021deep} for domain-aware adaptation in streaming-based applications. However, the external LM size is often limited to a range of $10$M to $100$M for on-device deployment. Given these limitations, cloud-based second-pass rescoring with LLMs may be a promising approach that leverages frozen pretrained models and leverages in-context learning. 

\begin{figure}[b!]
     \centering
     \includegraphics[width=.3\textwidth]{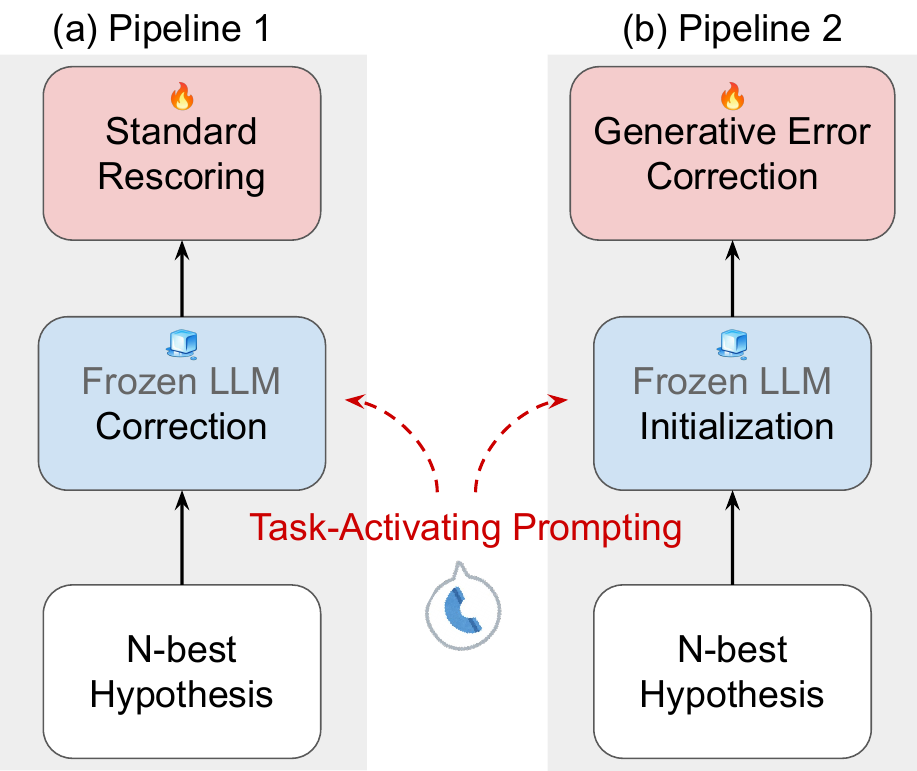}
     \caption{Two ASR post-processing frameworks using LLMs: (a) correct errors (e.g., grammar~\cite{guo2019spelling}) before applying a standard rescoring model, or (b) perform zero/few-shot rescoring; with optional task-activating prompting (Section~\ref{sec:tap}).}
     \label{fig:overview}
\end{figure}

Toward this end, in this work we explore novel ASR post-processing pipelines that utilize frozen LLMs by exploiting in-context learning.  We consider two ASR second-pass pipelines, as shown in Figure~\ref{fig:overview}:

{\bf $\mathcal{P}$ipeline 1:} a standard rescoring system takes in N-best output from a first ASR pass, and is trained to minimize the word error rate (MWER) by reranking the hypotheses. As illustrated in Figure~\ref{fig:overview}(a), an LLM in-context learning process is inserted into the pipeline to post-process first-pass hypotheses to apply error correction.

{\bf $\mathcal{P}$ipeline 2:} a new task-activating prompting method is used to initialize the frozen LLM with task-oriented instructions. A list of N-best ASR hypotheses is formatted as input to the LLM, thus allowing ``in-context learning initialization'' and/or ``in-domain fine-tuning'' (e.g., using adapters for parameter-efficient model update) that results in an improved speech transcription.   




In the remaining sections we present a first exploration of this novel way to utilize LLMs for the ASR task, demonstrate its surprising effectiveness, and compare results with different in-context learning schemes, as well as those of standard rescoring methods.

\section{Related Work}

\textbf{LLM-based post-processing to improve hypotheses.} 
Error correction post-processing~\cite{liao2020improving, yang2022asr} aims to fix grammar or deletion errors in output sentences and has been shown to improve the first-pass hypotheses generated from end-to-end ASR. A key characteristic of correction techniques is their reliance on pretrained LLMs, which benefit from rich contextual information. Liao \textit{et al.}~\cite{liao2020improving} propose ASR post-processing for readability, by extracting semantic expressions and generating readable text from ASR transcriptions.
N-best T5~\cite{ma2023n} used the T5 encoder-decoder architecture for rescoring with discriminative training.



\textbf{Zero-shot learning for acoustic and language modeling.}
Prior work has demonstrated that language modeling can generalize to zero-shot multi-tasks without exemplars \cite{wei2022finetuned, sanh2021multitask, zhong2021adapting}.
However, zero-shot and few-shot language modeling techniques often rely on the fine-tuning, which requires redeployment of pretrained models.



\textbf{In-context learning based on information prompting.}
In-context learning (ICL)~\cite{brown2020language, dai2023can} induces a single model to perform domain-agnostic inference without fine-tuning by providing a single or few prompts, thus addressing the aforementioned limitations. Prior study~\cite{min2022rethinking} has shown the ground truth demonstrations impose smaller effect than the author expected and significant zero-shot performance improvement under ICL framework. It implies external information gain can be extracted from frozen pretrained LLMs itself, if we select correct prompting strategy. However, ICL has its own shortcomings regarding tasks of reasoning. The chain-of-thought (CoT) prompting~\cite{weichain} decomposes the reasoning tasks by providing models with a sequence of questions or prompts that gradually guide the model to make predictions for a target task. While employment of CoT prompting is attributed to few-shot setup, LLMs has been proven to be the zero-shot reasoner given the single and specific prompt~\cite{kojimalarge}. In this work, we applied the above ICL techniques to ASR rescoring for the first time and empirically evaluate the their performance individually against the baseline.

\begin{figure}[tb]
     \centering
     \includegraphics[width=.48\textwidth]{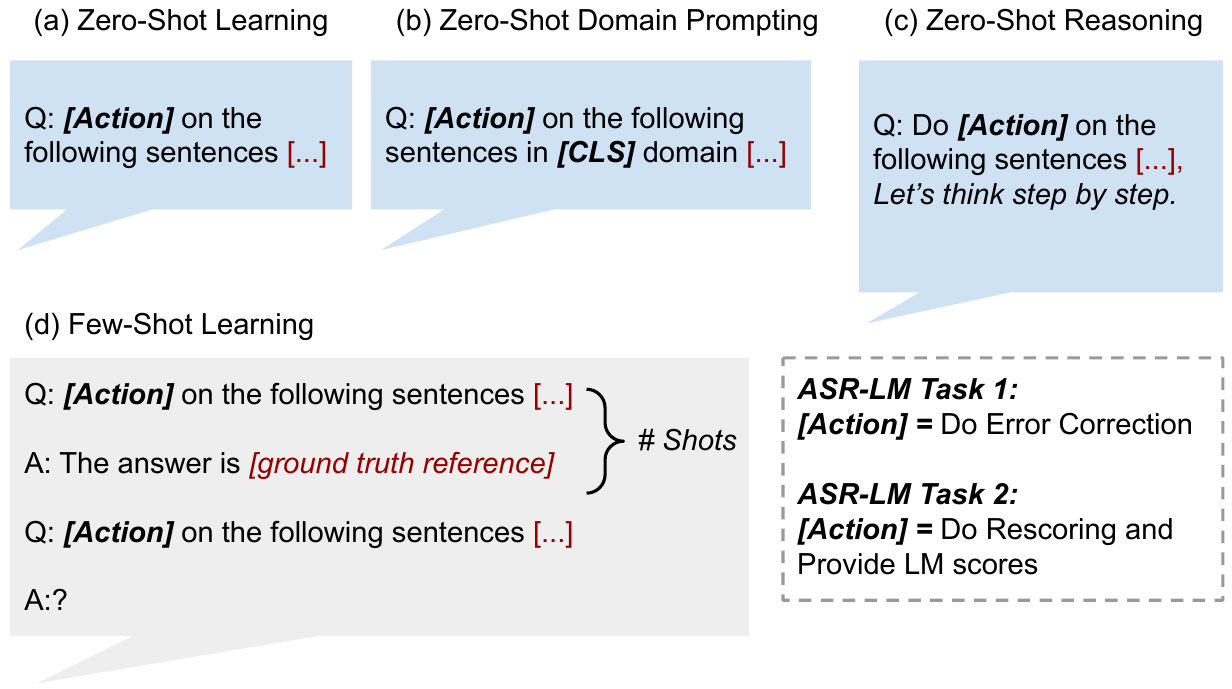}
     \caption{Four LLM in-context learning uses for ASR 2nd pass}
     \label{fig:icl:2}
\end{figure}

\section{Method}

We now review some recent advances in in-context learning techniques~\cite{weichain, brown2020language, kojimalarge} and describe how they can be incorporated into second-pass rescoring applications. 




\subsection{In-Context Learning Background and Techniques}
In-context learning~\cite{brown2020language} can emerge from modeling long-range coherence in the pretraining data. Based on a recent theoretical justification~\cite{xie2021explanation} by Bayesian inference, LLM would have implicitly learned to infer a latent concept during its pretraining stage. As an empirical result, in-context learning occurs if the LM can still infer the shared concept across examples (e.g., task instruction or prompts) to perform a target task. To model the in-context learning process, we can formulate its distribution over token $o$ within the vocabulary $O$ by sampling a latent \textit{confounding variable}~\cite{floccia2006does} $\theta$ from its population $\Theta$. 



The prediction over the pretraining distribution could be inferred by marginalizing over the confounding variable $\theta$: 
\begin{equation}
p_{\text{prompt}} = p(o_1,...o_T) = \int_{\theta \in \Theta}^{} p(o_1,...o_T|\theta)p(\theta)\,d\theta.
\end{equation}
Under the in-context learning framework, prediction sequence $O_i$ is inferred from the pretraining distribution conditioned on a prompt variable $\theta^{\ast}$, test-time sample (questions we would like to answer) $x_{\text{test}}$ and its in-context predictor $p_{\text{prompt}}(y|x)$:
\begin{equation}
    y_{\text{test}} \sim p_{\text{prompt}}(y|x_{\text{test}}, \theta^{\ast}).
    \label{eq:2}
\end{equation}

For instance, a simple prompt to empower in-context learning is to directly provide a ``\textit{task-oriented question}’’ to the pretrained LLM, as shown in Figure~\ref{fig:icl:2}(a). We further illustrate more  in-context learning setups in the following subsections.


\subsubsection{Zero-shot domain-hint prompting}
In the zero-shot setting, given a prompt template function $r()$ and $\theta^{\ast}$ as the domain-specific confounding variable (e.g., airline travel), a pretrained LLM models the conditional probability of the original input $x$ and target $y$, even if they were never trained, into their template function $r_{x}(x)$ and $r_{y}(y)$.
\begin{equation}
    r_{y}(y_{\text{test}}) \sim p_{\text{prompt}}(r_{y}(y)|r_{x}(x_{\text{test}}), \theta^{\ast}).
\end{equation}

In this work, we consider two general acoustic domains for making template function as a hard-coded input of ``\textit{airline information}'' or ``\textit{financial market}'', as shown in Figure~\ref{fig:icl:2}(b). 

\begin{figure*}[t]
\begin{center}
\parbox[t]{0.95\textwidth}{\footnotesize
\emph{\textbf{Q}: Do you know speech recognition?}\par
\emph{\textbf{R}: Yes, I am familiar with speech recognition. Speech recognition, also known as automatic speech recognition (ASR) or speech-to-text, is the process of converting spoken language into text. This technology involves using algorithms and machine learning models to analyze and transcribe the acoustic features of spoken words and phrases. Speech recognition has many applications, including voice-controlled assistants, automated phone systems, and transcription services.}
\par
\emph{\textbf{Q}: Do you know language model for speech recognition?}\par
\emph{\textbf{R}: Yes, I can provide you with some information about language models used in speech recognition. Language models are an essential component of Automatic Speech Recognition (ASR) systems. They help to predict the likelihood of a sequence of words, given the context. This helps the ASR system to better understand and transcribe spoken language into text. This separate model is typically more complex and powerful than the initial language model used for the transcription, and it is used to re-score the transcription based on the probability of the words occurring in the given context. The rescoring process involves taking the output of the initial language model, which is usually based on statistical methods such as Hidden Markov Models, and then applying a more advanced language model, such as a neural network-based language model, to generate a more accurate transcription. This is accomplished by re-ranking the possible transcriptions based on the probabilities assigned by the more advanced language model. Language model rescoring has been shown to significantly improve the accuracy of speech recognition systems, particularly in noisy or challenging environments where the initial language model may not perform well.}\par
\emph{\textbf{Q}: Could you give a possible example of language model rescoring with some hypotheses?} \par
\emph{\textbf{R}: Sure, here is an example of language model rescoring for ASR with 5-best hypotheses:} \par
\qquad \qquad \emph{1. recognize speech with artificial intelligence.} \par
\qquad \qquad \emph{2. recognized speech with artificial intelligence.} \par
\qquad \qquad \emph{3. recognize speech with artificial intelligent.} \par
\qquad \qquad \emph{4. reckon eyes speech with artificial intelligence.} \par
\qquad \qquad \emph{5. recognize peach with artificial intelligence.} \par
\quad \ \\ \emph{After rescoring, the model assigns a higher probability to the first hypothesis, which is the correct transcription of the input. The other hypotheses receive lower probabilities, reflecting their decreased likelihood given the improved language model. In this case, the top-ranked hypothesis after rescoring would be chosen as the final output: recognize speech with artificial intelligence} \par
\emph{\textbf{Q}: Nice job, I will provide some examples as a demonstration from [\textcolor{blue}{target domain}]. The 10-best hypothesis is:[\textcolor{blue}{hypotheses list from training set}], and I would expect your output is: [\textcolor{blue}{corresponding transcription}]. Following this example, could you report the true transcription from the following 10-best hypotheses:? [\textcolor{blue}{hypotheses list for inference}]}
}
\end{center}
\caption{Queries (Q) and responses (R) for N-best evaluation and correction by task-activating prompting (TAP) of LLMs}
\label{fig:tap}
\end{figure*}

\subsubsection{Zero-shot reasoning}
\label{sec:3:1:2}
Zero-shot reasoning~\cite{kojimalarge} employs chain-of-thought prompting~\cite{weichain} in a zero-shot setting with only two prompts: (i) reasoning extraction and (ii) answer extraction. Based on \cite{kojimalarge}, the reasoning extraction step uses a fixed and canonical prompt: \textit{Let's think step by step}, as shown in Figure~\ref{fig:icl:2}(c). 

In our experiments, we noticed that the reasoning extraction prompt is essential to boost the performance of zero-shot LLM rescoring. Under this zero-shot self-reasoning setup, the LLM output will first explain the task it is working on, then produce the actual task output. In the case of zero-shot rescoring, the LLM will first define a ASR-LM rescoring tasks and then provide LM scores for each N-best hypothesis.




\subsubsection{Few-shot and one-shot in-context learning}
A standard few-shot in-context learning process uses pairs of demonstrations~\cite{brown2020language} ``questions and targeted tasks,’’ retrieved from training data, to inform frozen LLMs for performing the target output, as illustrated in Figure~\ref{fig:icl:2}(d). One-shot in-context learning takes place by using a single demonstration as an input prompt for the frozen LLMs. Note that demonstrations (from an unseen training set) are distinct from test examples, and that the unsupervised-trained LLM has been reported to have a memory bottleneck based on term frequencies~\cite{razeghi2022impact}, which avoids potential data leakage issues for its few-shot learning evaluation reported in previous work~\cite{min2022rethinking, brown2020language, razeghi2022impact}.

\subsubsection{N-best hypotheses to transcription fine-tuning}

We introduce a hypotheses to transcription (H2T) mapping loss function: $\mathcal{L}_{\text{H2T}} = \sum_{i=1}^{N} - \{\log P(y^*| x_i, \mathbf{\Theta})+ \lambda \cdot \text{MSE}(s_i, P(y^*| x^{(i)}, \mathbf{\Theta}))\}$, where $P(y^*| x_i, \mathbf{\Theta})$ represents the probability of the true transcription ($y^*$) given the 
\textit{i}-th hypothesis ($x_i$) and the model parameters ($\mathbf{\Theta}$). 
To integrate acoustic information, a regularization term using mean squared error (MSE) is applied to penalize the model when there is a significant discrepancy between the predicted probabilities and the posterior probability scores with a $\lambda$ coefficient of $0.01$. 
 
Furthermore, we also consider parameter-efficient fine-tuning methods as in~\cite{chen2023hp}, which only update a small subset $\mathbf{\theta}\subset \mathbf{\Theta}$ of total trainable parameters to avoid potential overfitting that would hurt the generalization of the LLM~\cite{kumar2022fine, yang2021voice2series}.

\subsection{Task-activating Prompting (TAP) Framework}
    \label{sec:tap}

We now introduce a new in-context learning strategy that triggers the necessary sequential concepts for the ASR rescoring task, by utilizing multiple-round contextual sequences~\cite{hambardzumyan2021warp}. This technique is referred to as ``task-activating prompting'' (TAP). In this configuration, the LLM is given leading questions to clarify the task it needs to perform. Following this, the model is instructed to provide an example and, ultimately, it is presented with the top-N hypotheses from which to generate the actual output for the task. In our experiments, we noted that LLMs are capable of producing lists of the top N predictions, made up of utterances with analogous pronunciations. This demonstrates that LLMs assimilate acoustic (e.g., lattice-level) information during their pretraining phase. We illustrate the queries and responses used for task-activating prompting in Figure~\ref{fig:tap}.

We observed that the responses from InstructGPT vary slightly as a function of the random seed. In our experiments, we utilize the API key from OpenAI, and the prompts are fixed, except for the final query that contains the sample and test N-best lists. Recent work on ICL~\cite{ye2023explanation} focuses on the selection strategy for in-domain demonstration samples, which clearly could affect inference results. However, we leave demonstration selection for the N-best task to future work, and in our \emph{few-shot} learning experiments manually select longer demonstration utterances, following~\cite{min2022rethinking}.



\section{Experiments and Results}

We conduct our experiments by evaluating the final word error rate (WER) of a second-pass rescoring ASR system. First, we employ pretrained ASR models to decode audio from a target domain and collect the top-10 hypotheses. Then, we compare different in-context learning-based rescoring pipelines, including $\mathcal{P}_1$, the LLM-based error correction to enhance a standard ASR rescoring process, and $\mathcal{P}_2$, a prompting scheme that enables the LLM to directly perform ASR rescoring, as shown in Figure~\ref{fig:overview}.

\subsection{Pretrained ASR and Rescoring Model Training}

\label{section:4:1}
We select a conformer-based RNN-transducer~\cite{chiu2018state, gulati2020conformer} (760M parameters) pretrained with $960$ hours of Librispeech~\cite{panayotov2015librispeech}, 10k hours of the GigaSpeech~\cite{chen2021gigaspeech}, 24k hours of English data in VoxPopuli~\cite{wang2021voxpopuli}, and wav2vec2-based Libri-Light~\cite{kahn2020libri, baevski2020wav2vec} pretraining as the first-pass model to generate the hypothesis list. We follow the conformer architecture presented in~\cite{gulati2020conformer}. The pretrained model achieves a WER of $2.45$\% (test-clean) and $5.45$\% (test-other) on LibriSpeech without external LM. 

For $\mathcal{P}_1$, to obtain a standard neural LM for rescoring, we employ an ASR-LM training procedure similar to RescoreBERT~\cite{xu2022rescorebert}, which involves a two-stage process. The first stage is adaptation using a first-stage masked language model (MLM)~\cite{devlin2018bert} loss, and the second stage is to use minimum word error rate (MWER) training~\cite{prabhavalkar2018minimum, juang1997minimum, povey2005discriminative}. We choose the Albert~\cite{lan2021albert} \textit{base-v2} model with $4.8$ million trainable parameters, and apply rescoring~\cite{xu2022rescorebert} using the error-corrected hypotheses obtained with a frozen LLM.  

For $\mathcal{P}_2$, we examine how, by suitable prompt construction, the frozen pretrained LLMs can be instructed to perform ASR rescoring. The specific LLM configurations used are described next.

\subsection{Pretrained LLM Configurations}


\textbf{GPT-2} ($1.5$B): From the Generative Pretrained Transformer (GPT) family of causal models we selected this one~\cite{brown2020language, radford2019language} as our basic LLM for in-context prompting setup.
This version of GPT-2\footnote{The pretrained model is publicly accessible under MIT License \url{https://github.com/openai/gpt-2}}
is 100x smaller than very large teacher models, such as BLOOM~\cite{scao2022bloom}, making it much more suitable for real-world deployment. GPT-2 is trained primarily using Wikipedia~\cite{meritypointer} and Common Crawl~\cite{trinh2018simple}.

\textbf{OpenLLaMA} (13B): an open collection of transformer-based, decoder-only, causal language models ranging from 1B to 13B parameters \cite{openlm2023openllama}. It is trained exclusively on the RedPajama~\cite{together2023redpajama} dataset; we have confirmed that our Linguistic Data Consortium eval sets are not included.

\textbf{BLOOM} ($176$B): the first open-source LLM trained on the public supercomputer provided by the French government \cite{scao2022bloom}, BLOOM is available for reproducible studies of LLMs over $100$B. The model is pretrained using a large public collection of $498$ HuggingFace datasets~\cite{lhoest2021datasets}, comprising $1.61$ TB of text spanning $46$ natural languages and $13$ programming languages (our evaluation datasets are not included.)

\textbf{InstructGPT} ($175$B): an LLM created by training a GPT-3~\cite{brown2020language} model through reinforcement learning from human feedback (RLHF) \cite{ouyangtraining}. InstructGPT demonstrates improved zero-shot learning performance, benefiting from human knowledge transferred through RLHF, a process similar to student-teacher learning. InstructGPT is trained using human feedback without using open-domain data for evaluation.%
\footnote{InstructGPT is an earlier version of ChatGPT,. 
We did \textbf{not} use ChatGPT in our experiments due to its frequent revisions and unclear technical documentation.}


\subsection{Target-Domain Datasets}

We use the pretrained ASR model introduced in Section~\ref{section:4:1} to decode two public datasets, both in $\mathcal{P}_1$ (post-processing error correction) and $\mathcal{P}_2$ (pretrained LLM-based rescoring). 

\textbf{Airline Travel Information System (ATIS)}~\cite{hemphill1990atis} contains $4978$ training and $893$ utterances. ATIS comprises spoken queries for air travel information, such as flight times and availability. 

\textbf{Wall Street Journal (WSJ)}~\cite{marcus1993building} consists of transcribed audio recordings of read news articles from the Wall Street Journal, covering a wide range of topics and featuring a diverse set of speakers. We adapt the pretrained conformer model on the development set of \textit{train-si284} and test on \textit{93dev}.

\begin{table}[tb]
\caption{Oracle WERs for original and error-corrected N-best output, using $\mathcal{P}_1$ processing as shown in Figure~\ref{fig:overview}(a). The oracle error rates show the improvement in hypothesis quality as a result of post-processing using different sizes of LLMs.}
\label{tab:1:orcle}
\begin{adjustbox}{width=0.48\textwidth}
\centering
\begin{tabular}{lcc|c|c}
\toprule
$\mathcal{P}_1$: correction setup              & WSJ & ATIS  & WER$_\text{avg}$  & M-Size \\ \hline
(a) $N$-best        & 9.78 & 6.43 & 8.11 & -         \\ \hline
(a) + corrected by GPT-2       & 9.91 & 6.11 & 8.01 & 1.5B      \\
(a) + corrected by OpenLLaMA      & 9.95 & 5.73 & 7.43 & 13B      \\
(a) + corrected by BLOOM      & 9.21 & 5.64 & 7.42 & 176B      \\
(a) + corrected by InstructGPT  & \textbf{8.41} & \textbf{5.43} & \textbf{6.92} & 175B   
\\ \bottomrule
\end{tabular}
\end{adjustbox}
\end{table}

\begin{figure}[tb]
     \centering
     \includegraphics[width=.48\textwidth]{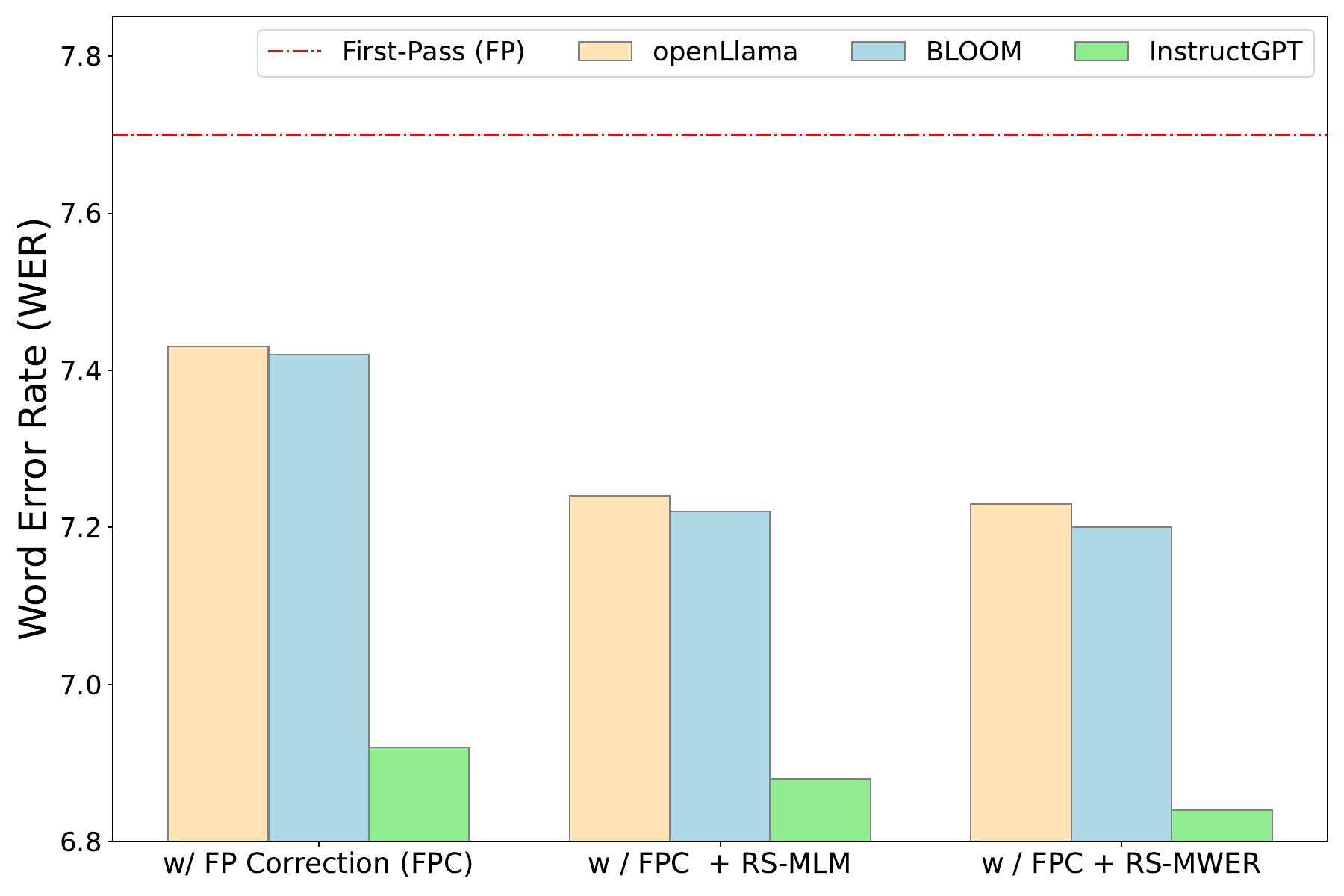}
     \vspace{-2em}
     \caption{$\mathcal{P}_1$ ASR rescoring (RS) training using hypotheses corrected by LLM. The dashed red line marks the $N$-best WER. The WER gradually decreases in the three stages of rescoring using our $\mathcal{P}_1$ processing: Stage 0, $N$-best hypothesis with LLM correction ($N$C); Stage 1, fine-tuned RescoreBERT~\cite{xu2022rescorebert} using the masked language modeling (MLM) loss; and Stage 2, MWER training. }
     \label{fig:incl}
\end{figure}
\subsection{$\mathcal{P}$ipeline 1 Results}

As shown in Table~\ref{tab:1:orcle}, we first use a pretrained LLM  for error correction, using the setup in~\cite{liao2020improving} to improve hypothesis quality as measured by oracle (minimum achievable) N-best error rates.  

Figure~\ref{fig:incl} shows how error correction by LLMs complements existing rescoring with adapted LMs, such as RescoreBERT~\cite{xu2022rescorebert}. We observe that the existing rescoring pipeline~\cite{xu2022rescorebert} reduces WER from $11.3\%$ to $8.7\%$ compared to its fine-tuning-only baseline. Furthermore, $\mathcal{P}$ipeline 1 employed the frozen pretrained language models to achieve an additional performance boost.



\begin{table*}[tb]
\caption{WERs on ATIS and WSJ using prompting variants to enhance the $\mathcal{P}_2$ in-context learning pipeline. We report the results of InstructGPT and BLOOM as LLMs over 100B; GPT-2 and OpenLLaMA do not perform consistently in this setting.}
\label{tab:in-context:e}
\centering
\begin{tabular}{lrr|rr}
\toprule
      & \multicolumn{2}{c}{\small WSJ} & \multicolumn{2}{|c}{\small ATIS} \\ \hline
In-context learning variant        & InstructGPT & BLOOM & InstructGPT & BLOOM \\
\hline
$\mathcal{P}_1$: LLM-corrected $N$-best w/ RescoreBERT~\cite{xu2022rescorebert}    & 10.13                            & 10.46  & 7.13                            & 8.46                         
\\ \hline
$\mathcal{P}_2$: (c) Zero-shot scoring  & 10.43                            & 11.23   & 7.95                            & 8.45                     \\
$\mathcal{P}_2$: (c) + zero-shot reasoning~\cite{kojimalarge} & 10.20                            & 11.88  & 7.77                            & 8.53                       \\
$\mathcal{P}_2$: (c) + domain-hint prompting~\cite{min2022rethinking} & 10.98                            & 11.45& 7.59                            &8.49                      \\\hline

$\mathcal{P}_2$: (d) Scoring with one example-pair  & 9.42                            & 9.45    & 6.34                            & 7.30                     \\ $\mathcal{P}_2$: (d) + zero-shot reasoning~\cite{kojimalarge} & 9.87                            & 11.46      & 7.25                            & 8.64                   \\
$\mathcal{P}_2$: (d) + domain-hint prompting~\cite{min2022rethinking} & 9.70                            & 10.99& 6.19                            & 7.12                      \\
$\mathcal{P}_2$: (d) + task-activating prompting (TAP) & \textbf{8.84}                           & \textbf{8.99} & \textbf{5.79}                            & \textbf{6.72}                      \\\bottomrule
\end{tabular}
\end{table*}

\begin{table}[tb]
\caption{WERs with $\mathcal{P}_2$ pipeline, using LLM in-context learning (ICL) for rescoring by zero-shot prompts illustrated in Figure~\ref{fig:icl:2}(a). Where indicated we use task-activating prompting (TAP) to condition the LLM.}
\label{tab:2:rescore:results}
\centering
\begin{tabular}{lcc}
\toprule
$\mathcal{P}_2$: zero-shot rescoring setup               & WSJ  & ATIS   \\ \hline
(a) Oracle            & 9.78  & 6.43       \\
(b) First pass        & 11.87 & 8.82     \\ \hline
(b) + $4$-gram LM        & 11.21 & 8.57      \\ \hline
(b) + frozen GPT-2                      & 29.56 & 27.51    \\
(b) + frozen GPT-2 \textit{w/ TAP}                             & 27.37 & 27.59     \\ \hline
(b) + frozen OpenLLaMA                       & 13.32 & 9.27     \\
(b) + frozen OpenLLaMA \textit{w/ TAP}                   & 11.53 & 8.61     \\ \hline
(b) + frozen BLOOM                        & 12.59 & 9.21     \\
(b) + frozen BLOOM \textit{w/ TAP}                   & \textbf{10.82} & \textbf{8.42}      \\ \hline
(b) + frozen InstructGPT                   & 9.97 & 7.15      \\ 
(b) + frozen InstructGPT \textit{w/ TAP}                       & \textbf{8.72} & \textbf{6.39}      \\ \bottomrule
\end{tabular}
\end{table}

\subsection{$\mathcal{P}$ipeline 2 Results}


\textbf{Case 1: Zero-shot learning.}
Table~\ref{tab:2:rescore:results} shows results for rescoring with in-context learning using different LLMs, as well as various baselines.
Note that for in-context rescoring, we extract LM scores from the model output responses, which differs from the standard approach of using the softmax outputs from the rescoring LM, for which we also report results.
The best $\mathcal{P}_2$ rescoring setup is the one using InstructGPT (Table~\ref{tab:2:rescore:results}, last row), achieving $19.7$\% relative WER reduction compared to rescoring with a fine-tuned GPT-2. Note that the frozen GPT-2 failed to give improvements over a 4-gram baseline, showing that a certain model size is required for generative error correction. For LLMs with over $100$ billion parameters, the use of prompting information showed better results compared to using the softmax scores directly. 

Next, we tested some popular prompting variants for in-context learning to possibly improve the performance of $\mathcal{P}_2$. As shown in Table~\ref{tab:in-context:e}, $\mathcal{P}_2$ and prompting with \textit{``step-by-step’’} (known as zero-shot reasoning in~\cite{kojimalarge}) achieved the best results for both LLMs (fourth row). It outperforms the one-shot learning variant (prompting with one input-output example; fifth row) by $1.8\%$ relative WER difference.
It is worth noting that the standard rescoring training $\mathcal{P}_1$ (first row) still outperforms zero-shot $\mathcal{P}_2$ by $14.6$\% relative.

\begin{figure}[bt]
     \centering
     \includegraphics[width=.48\textwidth]{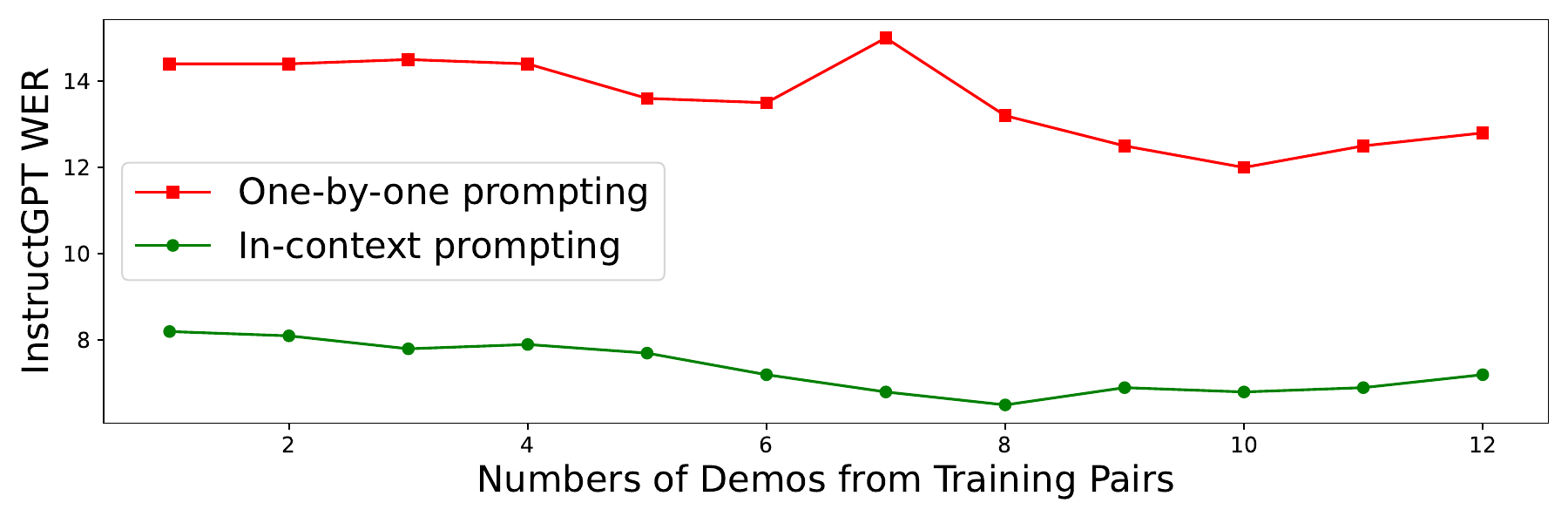}
     \vspace{-2em}
     \caption{WER results on ATIS and WSJ with few-shot learning based on InstructGPT, for increasing numbers of demonstration samples.
     ``One-by-one prompting'' resets the model history after each utterance, ``in-context prompting'' lets the history (and thus the examples provided) accumulate.}
     \label{fig:fs}
\end{figure}

\textbf{Case 2: Few-shot learning.} We gain some insight into the effects of in-context learning by considering few-shot learning in the form of conversational prompting. We feed InstructGPT examples drawn from the training portions of the two datasets, and report the results on the unseen test sets. As shown in Figure~\ref{fig:fs}, we see that frozen InstructGPT improved its rescoring performance as the number of training samples is increased from $1$ to $12$. It is better to let the model history accumulate (\textcolor{blue}{green} plot) than to reset it after each utterance (\textcolor{red}{red} plot), thereby compounding the effect of demonstrations.

\textbf{Case 3: In-domain fine-tuning.} We also obtained in-domain
fine-tuning results, where we use the training portions of all the speech
dataset to fine-tune the LLMs and then evaluate performance on the test sets; note that InstructGPT, being an API-only model, could not be fine-tuned.
For prompting we use TAP (Section~\ref{sec:tap}); however, we observed that after fine-tuning the exact method of prompting makes very little difference.
As shown in Table~\ref{tab:finetune}, fine-tuning with low-rank adapters (LoRA) outperforms full fine-tuning in the generative error correction case, as do residual adapters. One reason would be that adapters avoid modifying the parameters of a pretrained model (by inserting a neural module with a small number of additional trainable parameters that approximate the full parameter updates), allowing for efficient learning of the task without affecting the pretrained parameters of the LLM. \emph{LoRA}-based generative error correction introduces trainable low-rank decomposition matrices into the pretrained LLM layers, enabling the model to adapt to new data while keeping the original LLMs fixed to retain the pretrained knowledge. Specifically, LoRA performs a reparameterization of each model layer expressed as a matrix multiplication by inserting low-rank decomposition matrices. As a result, the representations (generated by the LLM) are not distorted due to task-specific tuning, while the adapter module acquires the ability to perform the error correction task.

Compared to previous state-of-of-the-art results with the Universal Speech Model (USM)~\cite{zhang2023google} with text injection training over 2B parameters, our best fine-tuning results improve upon their WSJ results (WER of 3.2\%).  This is noteworthy considering that our generative error correction method is based on a smaller underlying conformer-RNN-T ASR system.

\begin{table}[t!]
\caption{WERs on ATIS and WSJ, using fine-tuning (FT) and parameter-efficient adaptation to enhance the $\mathcal{P}_{2\text{-TAP}}$ pipeline}
\label{tab:finetune}
\begin{adjustbox}{width=0.48\textwidth}

\begin{tabular}{lrrr|rrr}
\toprule
                 & \multicolumn{3}{c}{WSJ}                         & \multicolumn{3}{|c}{ATIS} \\
Method        & \tiny GPT-2& \parbox{2em}{\tiny \centering Open-\\LLaMA} & \tiny BLOOM & \tiny GPT-2& \parbox{2em}{\tiny \centering Open-\\LLaMA} & \tiny BLOOM \\ \hline
FT + ranking~\cite{ma2023n}    & 9.93 & 8.09 & 8.42 & 6.34  & 3.71 & 3.75 \\ \hline
Full fine-tune   & 9.94 & 7.71 & 6.91 & 5.98 & 2.23 & 2.49 \\
Res.\ adapter~\cite{houlsby2019parameter}
                                        & \textbf{7.24} & 5.94 & 4.57        & \textbf{4.45} & 2.48 & 2.12 \\
LoRA~\cite{hu2021lora}
                                        & 7.52 & \textbf{2.11} &\textbf{2.81}& 4.57 & \textbf{1.69} & \textbf{1.97} \\
Prefix tuning~\cite{li2021prefix}
                                        & 9.32 & 6.99 & 7.43        & 5.32 & 2.63 & 2.74 \\
\bottomrule
\end{tabular}
\end{adjustbox}
\end{table}

Another parameter-efficient form of fine-tuning is prefix tuning \cite{li2021prefix}, where a continuous prompt prefix is inserted into the input and tuned to optimize task performance of the LLM.  However, this method gave worse results than the full or adapter-based fine-tuning methods for the larger LLMs.

\section{Conclusions}

We have explored how in-context learning can be applied to pretrained large language models for improving first-pass ASR N-best output {\em without fine-tuning}. For this task setting, we introduce two post-processing pipelines utilizing in-context learning. The first one uses a pretrained LLM for error correction prior to standard rescoring with a fine-tuned LM.  The second pipeline uses in-context learning by prompting, to instruct the frozen pretrained LLM to perform the rescoring task by itself. The latter method shows substantial gains over the first-pass ASR output and can be further enhanced with chain-of-thought and example prompting, as well as a new prompting scheme we call task-activating prompting. The best methods show 31\% (on ATIS) to 38\% (on WSJ) WER reduction over first-pass ASR, using a frozen InstructGPT, and better than with a fine-tuned GPT-2 LM.
Substantial additional gains are achieved by fine-tuning the LLM for the ASR output-correction task.  Post-processing with OpenLLaMA and LoRA fine-tuning achieves 86\% and 80\% WER reduction on ATIS and WSJ, respectively. These results are \textbf{below} the N-best oracle error rate, showing the LLM's ability to utilize prelearned knowledge to correct ASR output errors. Possible future work can look at how to integrate extra acoustic representations into pretrained LLMs for further enhancing generative ASR error correction. 




\clearpage
\small
\bibliographystyle{IEEEtran}
\bibliography{mybib}

\end{document}